# Combining Word and Character Vector Representation on Neural Machine Translation

Khaidzir Muhammad Shahih
PT Prosa Solusi Cerdas
Indonesia
khaidzir.shahih@prosa.ai

Ayu Purwarianti
School of Electrical Engineering and Informatics
Institut Teknologi Bandung
ayu@stei.itb.ac.id

*Abstract*— This paper describes combinations of word vector representation and character vector representation in English-Indonesian neural machine translation (NMT). Six configurations of NMT models were built with different input vector representations: word-based, combination of word and character representation using bidirectional LSTM (bi-LSTM), combination of word and character representation using CNN, combination of word and character representation by combining bi-LSTM and CNN by three different vector operations: addition, pointwise multiplication, and averaging. The experiment results showed that NMT models with concatenation of word and character representation obtained BLEU score higher than baseline model, ranging from 9.14 points to 11.65 points, for all models that combining both word and character representation, except the model that combining word and character representation using both bi-LSTM and CNN by addition operation. The highest BLEU score achieved was 42.48 compared to the 30.83 of the baseline model.

## I. INTRODUCTION

Neural machine translation (NMT) is machine translation that employs deep neural network as the translation algorithm. Here, the deep neural network topology used is usually the recurrent neural network (RNN) for both encoder and decoder. This topology is also known as sequence of sequence topology, where the input and output are usually a sequence of tokens. Compared to other deep neural network topology, RNNs are more suitable for processing natural languages because they are able to remember previous context. Various studies have been conducted for translation using RNN [1][2][3][4]. Before NMT era, the state of the art of machine translation is the phrase-based machine translation (PBMT). However, lately NMT has a performance that exceeds the performance of PBMT [5].

Most of the smallest units for RNN processing are in the form of words/character sets [1][6][2][7]. Another approach is to use characters as the smallest units (Zhang, Li, & Ji, 2016). Word-based usage generally has better performance than character-based [9].

Luong & Manning [4] combined representations of words and characters in NMT. Each in-vocabulary word is represented by an ordinary word-based vector representation, and each OOV word is processed per character so that the resulting representation of words is built on the constituent characters using bidirectional LSTM. The research showed this technique achieved higher BLEU score compared to NMT which uses only ordinary word representation.

Unlike Luong & Manning [4], Ma & Hovy [10] combined the ordinary word-based vector representation with word representation derived from its constituent characters using convolutional neural network (CNN) for POS tagger and named entity recognition (NER). The experimental results showed the technique had better accuracy for both tasks (POS tagger and NER) than the one with only word-based vector representation. The use of word and character representation can also be applied to the NMT case, because of the similar principle used: sequence to sequence learning with the encoder - decoder framework. The addition of character representation for NMT to word representation is expected to improve performance as for the POS tagger and NER cases.

## II. WORD AND CHARACTER REPRESENTATION FOR *NEURAL MACHINE TRANSLATION*

This section describes design of the neural machine translation with an encoder - decoder framework. In general, the process flow of the NMT is constructed as in Fig. 1. Input sentence as a sequence of words is then transformed into input layer of word embedding. In this research, several methods on combining the word-character vector representations as the input layer are compared. As a baseline, NMT with ordinary word-vector representation is chosen. The word embedding layer is inputted into the Encoder which output is inputted into the Decoder and resulted sequence of translated words.

Here, each unit of the RNN uses the LSTM model [11]. Bi-directional LSTM is used in both the encoder and the decoder [12]. To get better word order in the output, the attention mechanism [1] is applied.

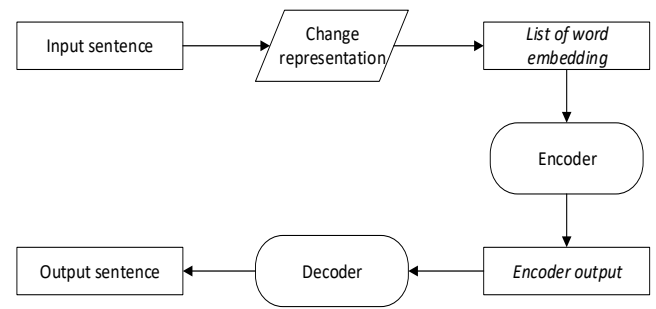

Fig. 1 General Flow of NMT Process

### A. NMT with Word Vector Representation (Baseline)

In NMT that is using only word vector representation, every word in input sentence is transformed into their word embedding vector, obtained from existing pre-trained word embedding. Preprocessing is done before every word is matched for their word embedding so that every word has the same format and consistent. The process flow from NMT based on ordinary word vector representation can be seen in Fig. 2.

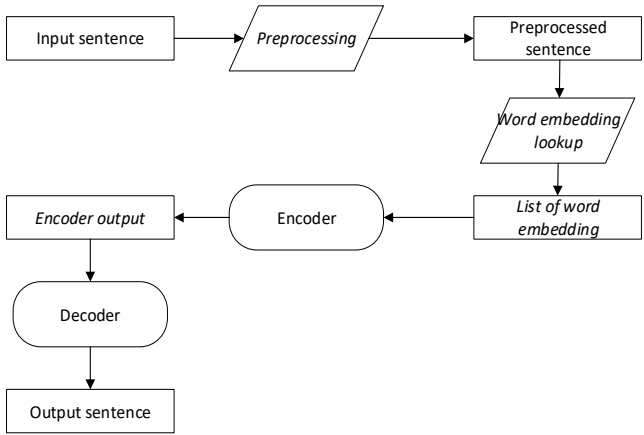

Fig. 2 Word-based NMT process flow

*B. NMT with Word and Character Vector Representation*

Here the input layer is built from ordinary word vector representation and character vector representation. The character vector representation is processed by using a deep neural network topology (bidirectional LSTM [7] or CNN model [10] [6]). Then this vector is concatenated with word vector that is obtained from pre-trained embedding. The concatenation technique is taken from [10] [6]. This combined vector is then forwarded to the NMT encoder. Architectural design for NMT using word and character vector representation can be seen in Fig. 3.

An embedding layer is used to change each character in a word to character embedding. The character embedding is then forwarded to bidirectional LSTM model or CNN model to get character-based representation from the input word. Fig. 4 shows an example of the word representation process of characters using bidirectional LSTM (bi-LSTM). Every char embedding *embedding* ($x_1$, $x_2$, $x_3$, dan $x_4$) is processed one by one in two directions, from left to right, and from right to left. Then the final vector representation is obtained by forward-hidden and hidden state backwards, i.e. $[\vec{h}_4^T; \overleftarrow{h}_1^T]^T$.

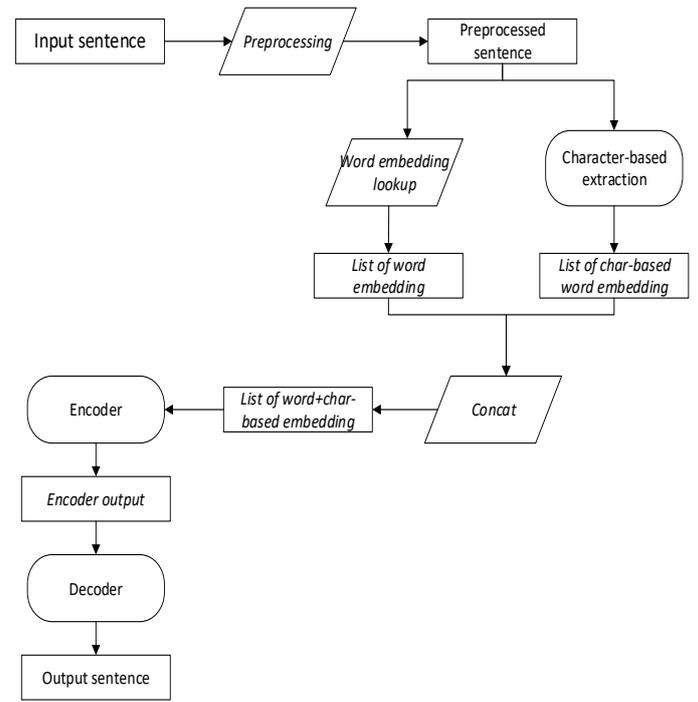

Fig. 3 Word and character based NMT process flow

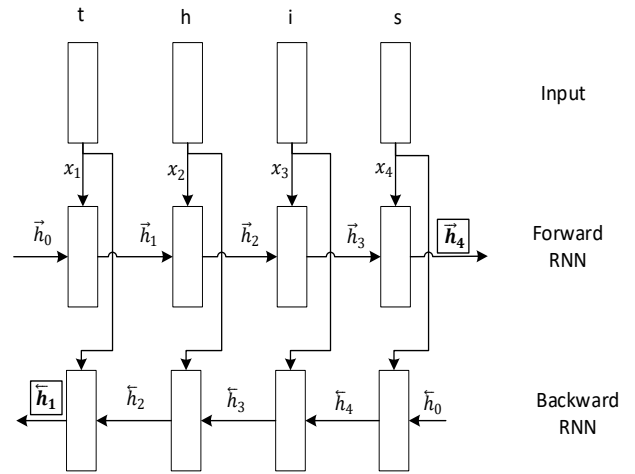

Fig. 4 Examples of character representation using BiLSTM

Figure 5 shows the character-based with CNN architecture. Formation of character-based representations is the same as that of Chiu & Nichols (2016). In the obtained matrix character embedding, convolution operations are performed using filters with windows 2, 3 and 4 for 1-max pooling operation to obtain the final vector representation of the processed word.

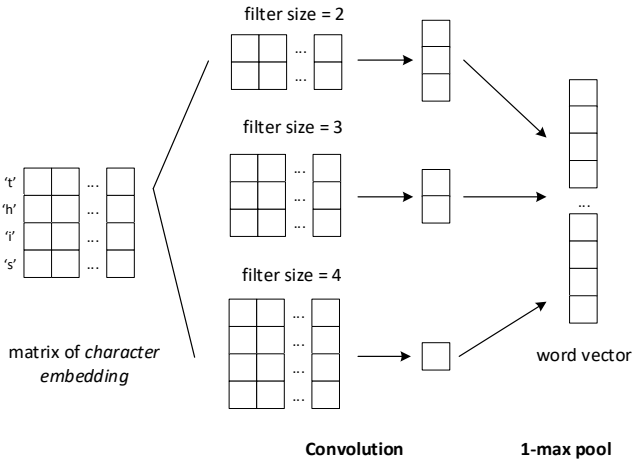

Fig. 5 CNN architecture to get character-based representation

In this research, a combination of character-based word representation will be carried out by the bidirectional LSTM and CNN models. Vector representation of the final character (vc) is obtained by operating a vector representation of the result of bidirectional LSTM ($v_{lstm}$) with a vector representation of the results of CNN ($v_{cnn}$). The operations tested for these two vectors include addition, average, and element-wise multiplication:

1. Addition : $v_c = v_{lstm} + v_{cnn}$
2. Averaging : $v_c = (v_{lstm} + v_{cnn})/2$
3. Multiplication : $v_c = v_{lstm} \odot v_{cnn}$

## III. Experiments

Each NMT is built using bidirectional LSTM for encoder and decoder. Attention mechanism is applied. Word embedding is obtained from Stanford GloVe [1] with 100 dimensions. The words in the pre-trained word embedding are all in lowercase.

The parameters used for each experiment are as follows.
1. Input size, the vector size used as encoder input: for the baseline model (word-based), input vector is obtained only from pre-trained word embedding used, that is 100 dimensions. For the other models, there is additional 100 dimensions obtained from character representation of word vector, resulting in 200 dimensions.
2. Hidden size (hidden state's vector size): 256 on encoder and 512 on decoder (256 dimensions from forward hidden state and another 256 from the backward).
3. Number of epochs: 100 epochs. The number of epochs that are limited to 100 and then the best model is taken.
4. Learning rate : 0.001

### A. Dataset

The dataset used in this experiment is an English - Indonesian pair news sentences which is translation from one another. The sentences in the dataset have been preprocessed. The preprocessing in step is lowercasing for each letter and do a tokenization in either an English sentence or an Indonesian sentence. The tokenization process is done by using NLTK[2] library.

The dataset used is divided into training data and test data. Training data is used to train the constructed neural machine translation model. The training data size is 10,000 pairs of sentences and the size of the test data is 1,000 sentences. TABLE 1 shows the statistics of the training data used. The testing metric used is the BLEU score [13].

TABLE 1
STATISTICS OF TRAINING DATA

| Statistic | Source language (English) | Target language (Indonesian) |
|---|---|---|
| Unique words | 972 | 907 |
| Minimum of sentence length | 1 | 1 |
| Maximum of sentence length | 45 | 45 |
| Average of sentence length | 9.26 | 7.92 |

### B. Scenarios

Six different configurations of NMT were tested on our experiments:
1. Word-based model (**word**) as baseline
2. Word and character based model using bi-LSTM (**wordchar_bilstm**)
3. Word and character based model using CNN (**wordchar_cnn**)
   The CNN model used was built using a 34 filter matrices with filter size of 2, 33 matrices with filter size of 3, and 33 matrices with filter size of 4. The ReLU function is performed. Then, 1-max pooling operation is performed. So the final vector has 100 dimensions, same as word embedding's dimension used.
4. Word and character based model by adding bidirectional LSTM and CNN (**wordchar_combine_add**)
5. Word and character based model by averaging bidirectional LSTM and CNN (**wordchar_combine_avg**)
6. Word and character based model by multiplicating bidirectional LSTM and CNN (**wordchar_combine_mul**)

The decoding process was done by using beam search technique with varying beam sizes up to 10.

---

[1] https://nlp.stanford.edu/projects/glove/

[2] www.nltk.org

## C. Results

TABLE 2 shows BLEU scores achieved by every model tested.

TABLE 2
BEST BLEU SCORES

| Model | Best BLEU | Nth-epoch | Decoding |
|---|---|---|---|
| *word (baseline)* | 30.83 | 40 | beam search (width=4) |
| *wordchar_bilstm* | 39.97 | 80 | beam search (width=4) |
| *wordchar_cnn* | 39.97 | 100 | beam search (width=3) |
| *wordchar_combine_add* | 30.75 | 60 | beam search (width=9) |
| *wordchar_combine_avg* | 41.49 | 95 | beam search (width=4) |
| *wordchar_combine_mul* | **42.48** | 80 | beam search (width=6) |

The test results show that the combination of word representation and character as input provides better results than just the use of word representation as input. This is indicated by BLEU score for all models with higher word and character representation compared to models that use only word representation from word embedding, except for the *wordchar_combine_add* model, by a margin, ranging from 9.14 to 11.65 points. The translation results of *wordchar_combine_add* model actually have a BLEU score lower than the baseline model.

TABLE 3
TRANSLATION EXAMPLES

| Input | it was n't the first to do this , but it was one of the best . |
|---|---|
| Reference | bukan yang pertama melakukan ini , tapi itu adalah salah satu yang terbaik . |
| *word (baseline)* | itu bukan pertama pertama pertama , tapi itu adalah satu yang terbaik musim ini . |
| *wordchar_bilstm* | itu bukan bagian pertama yang melakukan ini , itu salah salah satu yang terbaik . |
| *wordchar_cnn* | itu bukan yang pertama untuk melakukan , tapi itu bukan salah satu yang terbaik . |
| *wordchar_combine_add* | itu bukan yang pertama , tapi , tapi itu salah satu yang terbaik . |
| *wordchar_combine_avg* | itu bukan yang pertama yang melakukan ini , tapi itu salah satu yang terbaik . |
| *wordchar_combine_mul* | itu bukan yang pertama bagi itu , tapi itu salah satu yang terbaik . |
| *wordchar_combine_avg* | tidak ada yang datang untuk berbicara pada kami . |
| *wordchar_combine_mul* | tidak ada yang datang untuk berbicara dengan kita . |
| Input | the most recent was in april 2012 . |
| Reference | yang paling baru adalah pada bulan april 2012 . |
| *word* | yang ini adalah pada bulan oktober 2010 . |
| *wordchar_bilstm* | laporan besar tersebut pada pada 2011 . |
| *wordchar_cnn* | bagian yang paling ada pada bulan desember 2010 . |
| *wordchar_combine_add* | yang terbaru tahun ini pada bulan oktober . |
| *wordchar_combine_avg* | laporan tersebut adalah pada pada 2012 2010 . |
| *wordchar_combine_mul* | hal besar ini pada pada bulan april . |
| Input | i came to europe to play for chelsea and want to do it . |
| Reference | saya datang ke eropa untuk bermain untuk chelsea dan ingin melakukannya . |
| *word* | saya ingin ke sini untuk bermain dan saya ingin melakukan itu . |
| *wordchar_bilstm* | saya datang ke sini untuk bermain untuk sini dan ingin melakukannya dengan . |
| *wordchar_cnn* | saya datang ke chelsea untuk bermain di chelsea dan ingin melakukannya . |
| *wordchar_combine_add* | saya baru saja untuk untuk liverpool liverpool dan ingin melakukannya melakukannya . |
| *wordchar_combine_avg* | saya datang ke sini untuk untuk datang dan ingin melakukannya dengan itu . |
| *wordchar_combine_mul* | saya datang ke sini untuk bermain dan ingin ingin ingin melakukannya . |

TABLE 3 shows several examples of input sentences, references, and hypotheses from each NMT model for the case of sentences containing the words named entity or numeric words. It is seen that in some cases, the built NMT fails to translate several words which are named entities. The translation failure can be in the form of an untranslated named entity, such as the case of the word model for the 4th translation sentence (the word '*chelsea*' is not translated); and the case of translating a named entity that is translated into other different named entities, as in the case of the *wordchar_combine_add* model for the translation of the 4th sentence, word '*liverpool*' is shown instead of '*chelsea*'. Also, in 4th sentence, word '*april*' is not translated correctly by some models, instead of '*april*', word '*december*' or '*october*' are shown. One of another failure is translating numeric word, that can be seen in 3rd example, word '2010' is not translated correctly. Those failures is mainly due to the small number of dataset used.

## IV. CONCLUSIONS

The use of character vector representation in addition to word level vector representation in NMT has been shown to improve BLEU score compared to the word-based only model. The improvement of the BLEU score obtained by our experiments is in the range of 9.14 to 11.65 points. This results consistent to all tested models and scenarios, except for the model that combining bi-LSTM and CNN character representations using addition operation. Current work is mainly to show that incorporating character-level representation along with word-level representation of embedding can improve translation quality. The future works that can be done is to increase the size of dataset and add more layers on encoder to see whether translation quality is improved or not.